\documentclass[runningheads]{llncs}

\usepackage{eccv}

\usepackage{eccvabbrv}

\usepackage{graphicx}
\usepackage{booktabs}
\usepackage{marvosym}

\usepackage[accsupp]{axessibility}  

\usepackage{hyperref}

\usepackage{orcidlink}

\begin{document}

\title{Data-Efficient Generation for Dataset Distillation} 

\titlerunning{Data-Efficient Generation for Dataset  Distillation}

\author{Zhe Li\inst{1}\(^{(\textrm{\Letter})}\)\orcidlink{0009-0003-3101-718X} \and
Weitong Zhang\inst{2,3}
\and
Sarah Cechnicka\inst{2,3} \orcidlink{0009-0008-3449-9379} 
\and
Bernhard Kainz \inst{1,3}\orcidlink{0000-0002-7813-5023}}

\authorrunning{Z.~Li et al.}

\institute{Image Data Exploration and Analysis Lab, Friedrich-Alexander-Universität Erlangen-Nürnberg, Germany \\
\email{zhe.li@fau.de}\\
\and
UKRI Centre for Doctoral Training in AI for Healthcare, \\
Imperial College London, UK  \\
\and
Department of Computing, Imperial College London, UK \\
}

\maketitle

\begin{abstract}
  While deep learning techniques have proven successful in image-related tasks, the exponentially increased data storage and computation costs become a significant challenge. Dataset distillation addresses these challenges by synthesizing only a few images for each class that encapsulate all essential information. Most current methods focus on matching. The problems lie in the synthetic images not being human-readable and the dataset performance being insufficient for downstream learning tasks. Moreover, the distillation time can quickly get out of bounds when the number of synthetic images per class increases even slightly. To address this, we train a class conditional latent diffusion model capable of generating realistic synthetic images with labels. The sampling time can be reduced to several tens of images per seconds. We demonstrate that models can be effectively trained using only a small set of synthetic images and evaluated on a large real test set. Our approach achieved rank \(1\) in The First Dataset Distillation Challenge at ECCV 2024 on the CIFAR100 and TinyImageNet datasets.
\end{abstract}

\section{Introduction}
\label{sec:intro}

As deep learning methods become more established in computer vision, the dataset sizes and numbers of model parameters grow exponentially for various tasks such as image classification.
These ever-growing datasets require increased memory for storage and incur higher computational costs during network training.
To address this challenge, researchers started to focus on dataset distillation.
Dataset distillation aims to create a smaller, synthetic dataset that captures the essential information from a large dataset. In addition to reducing storage and computation costs, dataset distillation can anonymize individual information effectively, thus enhancing personal privacy protection.

\begin{figure}[t]
    \centering
    \includegraphics[width=\linewidth]{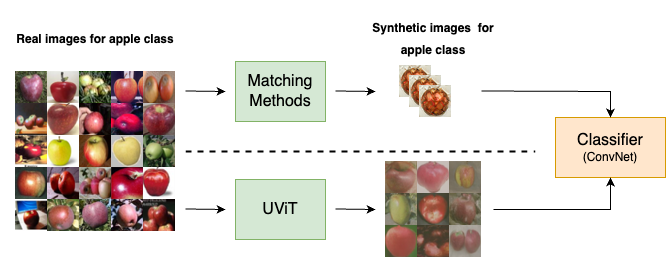}
    \caption{An overview of dataset distillation task. The upper half of the image represents the typical matching methods and the lower half demonstrates how to generate synthetic images using a diffusion model.}
    \label{fig:teaser}
    \vspace{-3mm}
\end{figure}

Most existing dataset distillation approaches~\cite{zhao2020dataset,zhao2023DM,cazenavette2022dataset,wang2022cafe,sajedi2023datadam,liu2022dataset,du2023minimizing,zhu2023rethinking} focused on various matching algorithms and achieved reasonable performance as can be seen in the upper half of Fig~\ref{fig:teaser}. The computation cost may increase exponentially when the number of images per class increase.
Furthermore, the quality of synthetic images is often not human-readable.
The challenges of reducing computational costs and increasing the image quality are still significant.

To address the problem of generating human-readable distilled synthetic images, we propose using class conditional diffusion models as can be seen in the lower half of Fig~\ref{fig:teaser}. Li et al.~\cite{li2024image} applies a diffusion model for dataset distillation of medical images. We propose that this approach could also be beneficial when training on natural images. The quality of these synthetic images relies on the effectiveness of diffusion models. One positive aspect is that many diffusion models can generate realistic synthetic images, such as UViT~\cite{bao2023all} and GAN-based models~\cite{sauer2022stylegan}. 
Besides generating images with good quality in different class, the diffusion models are capable of capturing the image distribution of the training dataset. Also, we only need to train the diffusion models once, then generate as many synthetic images as needed in advance. This property can significantly reduce the computational cost.

In this paper, we train a UViT~\cite{bao2023all} on the large real image datasets because it can be used as a backbone model and reports low Fréchet Image Distance (FID)~\cite{heusel2017gans} compared to other class conditional image diffusion models. 
In typical settings, the distilled small dataset contains only a few images per class (IPC), for example, 5 or 10 images. In this context, the small synthetic dataset may not accurately capture the entire data distribution and may not fully represent the original dataset due to information leakage.
We generate as many synthetic images as possible within ten minutes (the requirement of the Challenge) to comprehensively cover the data distribution. Because the synthetic images are generated in advance before training the classifier, the distillation time increases linearly with the number of images. The time can be reduced to generate several tens of images per second, so the total time remains reasonable even when generating a large synthetic dataset.

Our results are the rank \(1\) in Track \(2\) of The First Dataset Distillation Challenge at ECCV 2024 and outperform the competition with a huge margin on both CIFAR100~\cite{krizhevsky2009learning} and TinyImageNet~\cite{deng2009imagenet} datasets.

Our contribution consists of:
\begin{enumerate}
  \item We propose using a class conditional diffusion model for dataset distillation to distill synthetic images from large real datasets.
  \item We synthesize human readable images with our framework. The class conditional diffusion model has the ability to generate enough synthetic images in short time and represent the dataset distribution. It reduces the computational cost without an exponential increase with regard to the number of synthetic images.
  
  \item In our experiments, we evaluate our approach on two datasets CIFAR100~\cite{krizhevsky2009learning} and TinyImageNet~\cite{deng2009imagenet}. We report the average results of test accuracy of three runs.
\end{enumerate}

\section{Related Work}
\label{sec:related}

This field started from model selection~\cite{wang2018dataset, such2020generative}, then diversified into various matching methods to align the real and the synthesized data, such as Dataset Condensation with Gradient Matching (DC)~\cite{zhao2020dataset,zhao2021dataset,zhang2023accelerating}, Distribution Matching (DM)~\cite{zhao2023DM,zhao2023improved}, Matching Training Trajectories (MTT)~\cite{cazenavette2022dataset}, and Sequential Subset Matching~\cite{du2024sequential} which update different level features sequentially.
Beyond matching strategies, methods to align features of convolutional networks~\cite{wang2022cafe,sajedi2023datadam} have been proposed to improve performance.
Moreover, factorization~\cite{liu2022dataset}, accumulated trajectory errors~\cite{du2023minimizing}, calibration techniques~\cite{zhu2023rethinking}, and Frequency Domain-based approach~\cite{shin2024frequency} are also proposed.

Recognizing the limitations of pixel space, characterized by high-frequency noise, ~\cite{zhao2022synthesizing} shifted focus to synthesizing images in the latent space using pre-trained GANs, thereby extracting more informative samples. 
Aiming for simplicity and efficiency, ~\cite{cazenavette2023generalizing} utilized a pre-trained StyleGAN-XL~\cite{sauer2022stylegan} to create a single synthetic image per class from latent space, streamlining the distillation process from real datasets.
The field continues to evolve with methods addressing various phases of dataset distillation, including the introduction of the distillation space concept~\cite{liu2023fewshot}, the implementation of a clustering process for selecting mini-batch real images~\cite{liu2023dream}, and new matching metric with mutual information~\cite{shang2024mim4dd}.
~\cite{liu2024mgdd} adopted an encoder-decoder model to generate synthetic images in a forward in meta-learning. ~\cite{gu2023efficient} fine-tuned a diffusion model to generate realistic images.

\section{Preliminaries}
\label{sec:preliminaries}

\subsection{Data Distillation}

Data distillation is a process aimed at reducing the size of a training dataset \(\mathcal{T}\) while preserving its fundamental information for effective model training. Particularly in scenarios where computational resources are limited, or where privacy concerns necessitate the minimization of data usage. The goal is to create a distilled dataset \(\mathcal{S}\) that is significantly smaller than the original dataset but still allows a model \( f_\theta \) trained on it to perform comparably to a model trained on the full dataset. Following~\cite{sachdeva2023data}, we can define this formally as follows:

\begin{definition}[Data Distillation]
Given a model class \( \mathcal{F} \) and a training set \(\mathcal{T} =\{(x_i, y_i)\}_{i=1}^{N}\) where \(x_i\in \mathbb{R}^{3 \times H \times W}\) denotes an image, \(y_i \in \{0, 1, 2, ..., C\}\) represents its corresponding class label, in a dataset with \(C\) total classes and \(N\) total samples. The objective of data distillation is to find a distilled set \(\mathcal{S}=\{(s_i, y_i)\}_{i=1}^{M}\) of size \( M \ll N \) such that the model \( f_\theta \in \mathcal{F} \) trained on \( \mathcal{S} \) approximates the performance of the same model trained on \( \mathcal{T} \):

\begin{equation}
\mathcal{S} = \underset{\mathcal{S}'}{\arg \min} \ \mathbb{E}_{f_\theta \in \mathcal{F}}\left[ \mathcal{L}(f_\theta(\mathcal{T}_\text{test})) \mid \mathcal{S}' \right],
\end{equation}

where \( \mathcal{L}(\cdot) \) represents the loss function, \( \mathcal{T}_\text{test} \) is the test set, and \( f_\theta(\mathcal{S}') \) is the model trained on the distilled dataset \( \mathcal{S}' \).
\end{definition}

Often, one needs to consider both the informativeness and the diversity of the data points in \( \mathcal{S} \). A systematic strategy to data distillation also involves ensuring that the distilled dataset effectively captures the underlying data distribution of \( \mathcal{T} \), which is crucial for the model's generalization capabilities.

\begin{definition}[Optimal Data Distillation]
The optimal data distillation can be described as finding a subset \( \mathcal{S}  \) such that the risk \( R(f_\theta) \) on the true data distribution \( \mathcal{D} \) is minimized. 

\begin{equation}
\mathcal{S} = \underset{\mathcal{S}'}{\arg \min} \ R(f_\theta) = \underset{\mathcal{S}'}{\arg \min} \ \mathbb{E}_{\mathbf{x} \sim \mathcal{D}} \left[ \mathcal{L}(f_\theta(\mathbf{x})) \right],
\end{equation}

where \( R(f_\theta) \) is the risk function, and \( \mathcal{D} \) represents the true underlying data distribution.
\end{definition}

The challenge lies in selecting or generating a distilled dataset that retains the key characteristics of the original dataset necessary for effective learning. The distilled dataset \( \mathcal{S} \) must effectively minimize the empirical risk \( R(f_\theta) \) by preserving both the informativeness and diversity of the original dataset \( \mathcal{T} \).

\subsection{Generative Diffusion Models and Risk Minimization}

Generative diffusion models can address the risk minimization problem. Specifically, these models can generate synthetic data points \( \mathbf{x}_\text{gen} \) that are drawn from a distribution \( \hat{\mathcal{D}} \) designed to approximate the true data distribution \( \mathcal{D} \). This synthetic data can be used to enhance the distilled dataset \( \mathcal{S} \), thereby reducing the risk \( R(f_\theta) \). Using generative models, we redefine the risk function as follows:

\begin{equation}
\hat{R}(f_\theta) = \mathbb{E}_{\mathbf{s} \sim \hat{\mathcal{D}}} \left[ \mathcal{L}(f_\theta(\mathbf{s})) \right],
\end{equation}

where \( \hat{\mathcal{D}} \) represents the generative diffusion distribution of data. In this approach, \( \hat{\mathcal{D}} \) can be specifically tailored to capture the most informative and diverse aspects of \( \mathcal{D} \), thus effectively contributing to distillation. Furthermore, the generative diffusion model reduces the empirical risk by assuring that the distilled dataset is representative of the original distribution, in addition to serving as a tool for data enhancement.

\subsection{Problem Formulation}

This report focuses on developing and evaluating a diffusion-based generative model designed specifically for data distillation tasks. The objective is to construct a model \( p_\theta(\mathbf{x}_0) \) that can efficiently generate high-fidelity synthetic images, thereby creating a compact and highly informative distilled dataset.

The problem is formally defined as follows: Given a large dataset \( \mathcal{T} \), the goal is to generate a distilled dataset \( \mathcal{S} \) using a diffusion model such that the expected loss on a test dataset \( \mathcal{T}_\text{test} \) is minimized. The goal is to balance the trade-off between dataset size and model performance, ensuring that the distilled dataset is efficient for training while still maintaining or improving generalization capabilities.

\section{Method}
\label{sec:method}

\begin{figure}[t]
    \centering
    \includegraphics[width=\linewidth]{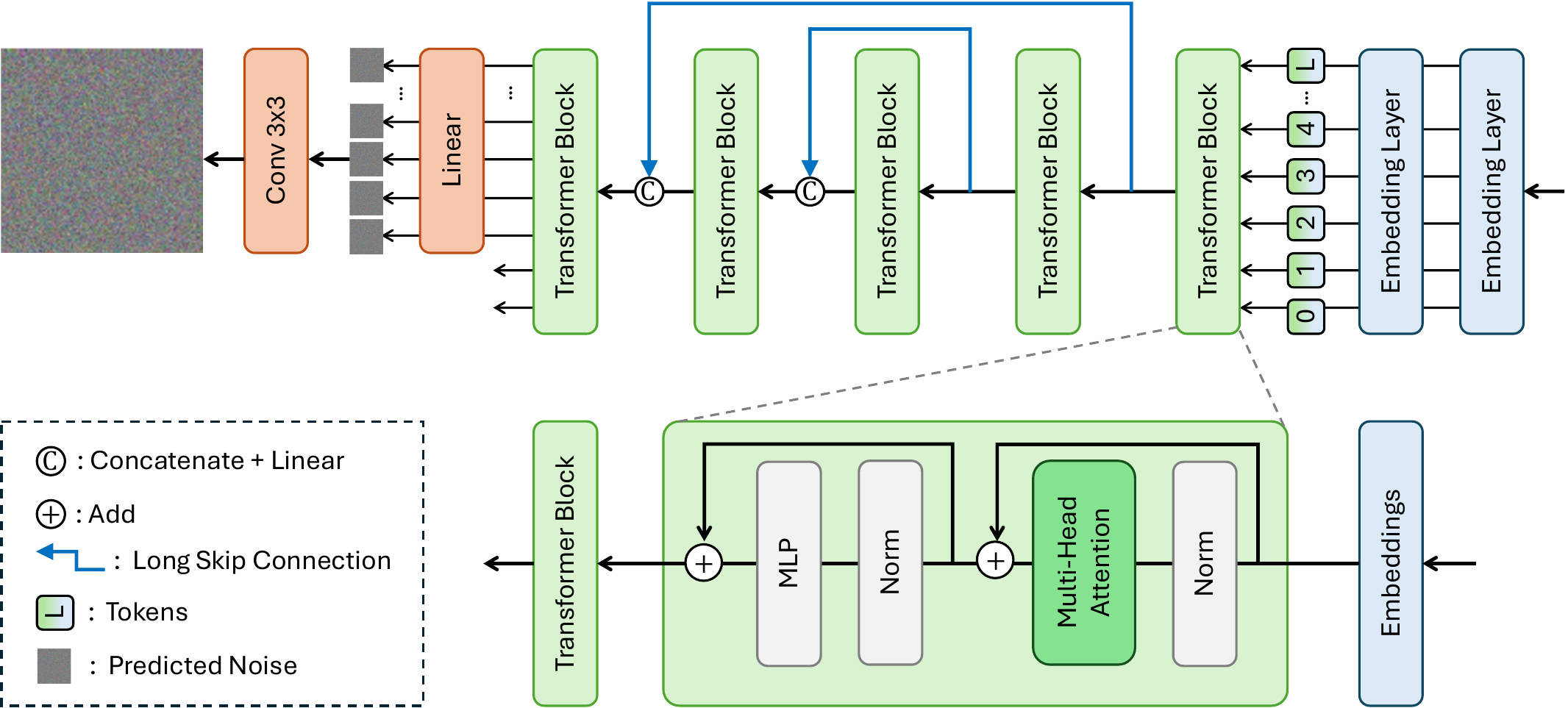}
    \caption{The architecture applied in diffusion models. The architecture iteratively refines noise predictions using a combination of multi-layer networks and multi-head attention mechanisms.}
    \label{fig:method}
    \vspace{-3mm}
\end{figure}

This section describes the architecture and mechanisms of the UViT model, specifically tailored for data distillation applications. The UViT framework integrates a Vision Transformer (ViT) backbone within a U-Net structure to effectively model complex data distributions in the latent space, thereby facilitating the generation of high-fidelity synthetic data that preserves essential information from the original dataset \( \mathcal{T} \).

\subsection{UViT Architecture}

The UViT model~\cite{bao2023all} is designed to leverage the strengths of both Vision Transformers and U-Net architectures~\cite{ronneberger2015u}, providing a robust backbone for diffusion-based generative tasks as shown in Figure~\ref{fig:method}. Our UViT consists of the following components:

\textbf{Vision Transformer Backbone.} The Vision Transformer (ViT) serves as the feature extractor within UViT. Unlike traditional CNNs, ViTs operate on image patches rather than individual pixels, capturing long-range dependencies and global contextual information effectively. In UViT, the image is first divided into non-overlapping patches, each of which is linearly embedded into a fixed-dimensional vector, each of size \( p \times p \). The sequence of embeddings \( \mathbf{z}_0 \) is processed by a series of Transformer layers, which include multi-head self-attention mechanisms and feed-forward networks. The patch embeddings are defined as:

\begin{equation}
\mathbf{z}_0 = [\mathbf{x}^1\mathbf{E}; \mathbf{x}^2\mathbf{E}; \dots; \mathbf{x}^N\mathbf{E}] + \mathbf{E}_{pos},
\end{equation}

where \( \mathbf{x}^i \) represents the \( i \)-th image patch, \( \mathbf{E} \) is the embedding matrix, and \( \mathbf{E}_{pos} \) denotes the positional embeddings added to retain spatial information. These patch embeddings are then fed into the Transformer encoder, which applies a series of multi-head self-attention layers and feed-forward networks to learn rich feature representations. The self-attention mechanism is designed as follows:

\begin{equation}
\text{Attention}(Q, K, V) = \text{softmax}\left(\frac{QK^T}{\sqrt{D}}\right)V,
\end{equation}

where \( Q \), \( K \), and \( V \) are the query, key, and value matrices derived from the input embeddings. 

\textbf{U-Net Structure.} UViT incorporates a U-Net architecture around the ViT backbone to facilitate efficient diffusion in the latent space. A bottleneck layer connects the encoder and decoder in a U-Net structure. The encoder consists of multiple downsampling layers, where the resolution of the feature maps is progressively reduced while increasing the depth of feature representations: 

\begin{equation}
\mathbf{h}^{(l)} = \text{Downsample}\left(\text{Activation}\left(\mathbf{W}^{(l)} \mathbf{h}^{(l-1)} + \mathbf{b}^{(l)}\right)\right),
\end{equation}

where \( \mathbf{h}^{(l)} \) is the output at layer \( l \), \( \mathbf{W}^{(l)} \) and \( \mathbf{b}^{(l)} \) are the weight and bias parameters, and the downsampling reduces the resolution. The decoder upsamples the feature maps back to the original resolution, reconstructing the input while incorporating contextual information:

\begin{equation}
\mathbf{g}^{(l)} = \text{Upsample}\left(\text{Activation}\left(\mathbf{W}^{(l)} \mathbf{g}^{(l+1)} + \mathbf{b}^{(l)}\right)\right) + \mathbf{h}^{(L-l)},
\end{equation}

where \( \mathbf{g}^{(l)} \) is the decoder output at layer \( l \), and \( \mathbf{h}^{(L-l)} \) represents the skip connection from the corresponding encoder layer \( L-l \). During the decoding process, these connections allow the model to retain high-resolution features from the encoder layers. This design maintains the structural integrity as well as fine details of the original data during data distillation tasks.

\subsection{Generative Diffusion Process}

The generative diffusion process in our UViT model is defined by two principal phases for data distillation: the forward diffusion process and the backward denoising process.

\textbf{Forward Diffusion Process.}
Our model begins with an original image \( \mathbf{x}_0 \), which is gradually perturbed by the addition of Gaussian noise across a series of time steps. This stochastic process can be described by a sequence of latent variables \( \mathbf{x}_1, \mathbf{x}_2, \dots, \mathbf{x}_T \), where each \( \mathbf{x}_t \) is conditionally dependent on \( \mathbf{x}_{t-1} \) through a Gaussian transition:

\begin{equation}
q(\mathbf{x}_t | \mathbf{x}_{t-1}) = \mathcal{N}(\mathbf{x}_t; \sqrt{\alpha_t} \mathbf{x}_{t-1}, (1-\alpha_t)\mathbf{I}),
\end{equation}

Here, \( \alpha_t \) is a time-dependent scaling factor that controls the variance of the added noise, and \( \mathbf{I} \) denotes the identity matrix. As \( t \) increases, \( \mathbf{x}_t \) becomes progressively noisier, approaching a state of near-complete noise at \( t = T \). This process effectively diffuses the information contained in the original image into a high-dimensional latent space, making it feasible for the synthetic representative data generation.

\textbf{Backward Denoising Process.}
Backward diffusion reverses the corruption introduced by forward diffusion, reconstructing the image \( \hat{\mathbf{x}}_0 \) from the noisy latent state \( \mathbf{x}_T \). For this reconstruction step, the conditional probability is:

\begin{equation}
p_\theta(\mathbf{x}_{t-1} | \mathbf{x}_t) = \mathcal{N}(\mathbf{x}_{t-1}; \mu_\theta(\mathbf{x}_t, t), \Sigma_\theta(\mathbf{x}_t, t)),
\end{equation}

where \( \mu_\theta(\mathbf{x}_t, t) \) represents the predicted mean, and \( \Sigma_\theta(\mathbf{x}_t, t) \) represents the predicted variance of the denoised image. The mean prediction is given by:

\begin{equation}
\mu_\theta(\mathbf{x}_t, t) = \frac{1}{\sqrt{\alpha_t}} \left(\mathbf{x}_t - \frac{\beta_t}{\sqrt{1 - \overline{\alpha}_t}} \epsilon_\theta(\mathbf{x}_t, t)\right),
\end{equation}

where \( \beta_t \) is the noise variance, and \( \epsilon_\theta(\mathbf{x}_t, t) \) is the noise estimate generated by the UViT model. By iterating over the image, the backward process refines it to approximate the original input, thereby generating synthetic images \( \mathbf{x}_\text{gen} \) that maintain the statistical properties of the training data while ensuring that no specific information is preserved.

\textbf{Diffusion with Data Distillation.}
The synthetic images \( \mathbf{x}_\text{gen} \) generated through this diffusion process are then used to construct the distilled dataset \( \mathcal{S} \). A data distillation creates a subset \( \mathcal{S} \subset \mathbf{x}_\text{gen} \) that is significantly smaller than the original dataset \( \mathcal{T} \) but retains its essential characteristics for effective model training. The selection of this subset is driven by the need to minimize the empirical risk \( R(f_\theta) \) on the true data distribution \( \mathcal{D} \), ensuring that the distilled dataset is informative as well as diverse in the track.

\section{Experiments}
\label{sec:experiments}

\subsection{Datasets} 

We apply our framework on two datasets: CIFAR100~\cite{krizhevsky2009learning} and TinyImageNet~\cite{deng2009imagenet}. CIFAR100~\cite{krizhevsky2009learning} consists of \(60,000\) images at the resolution \(32\times 32\). The train set consists of \(100\) image classes with \(500\) images per class. The test set has \(10,000\) images.
TinyImageNet~\cite{deng2009imagenet} consists of \(120,000\) images at the resolution \(64\times 64\). In the train set, there are \(200\) classes in total with \(500\) images per class. The validation set and test set have \(10,000\) images respectively.

\begin{table}
  \caption{The diffusion model configurations.}
  \label{uvitconfig}
  \centering
  \resizebox{0.8\columnwidth}{!}{%
  \begin{tabular}{lcccccccc}
    \toprule
    Models & Layers & Hidden Size & MLP Size & Head & Params \\
    \midrule
    UViT-Small & 13 & 512 & 2048 & 8 & 44M \\
    UViT-Mid & 17 & 768 & 3072 & 12 & 131M \\
    \bottomrule
  \end{tabular}
  }
\end{table}

\subsection{Implement Details}

In training, we use the AdamW optimizer~\cite{loshchilov2019decoupled} with a weight decay of 0.03 for all datasets and betas are set to \((0.99, 0.999)\). The learning rate is \(0.0002\). The batch size is \(1024\).
In inference, we sample \(50,000\) images and calculate the FID every \(500\) epochs.
When training on the CIFAR100 dataset, we initialize the small UViT~\cite{bao2023all} model by the pretrained CIFAR10 weights excluding the final label layer and finetuning the model for \(80,000\) epochs.
When training on the TinyImageNet dataset, we initialize the middle UViT~\cite{bao2023all} model by the pretrained ImageNet weights excluding the final label layer and finetuning the model for \(117,000\) epoches. We do not use AutoEncoder to generate latent features because the image size is so small.
The diffusion model configurations are listed in Table~\ref{uvitconfig}

The model is trained on 4 A100 GPUs. After training, we sample synthetic images with labels for \(10\) minutes. Then, we train the classifier ConvNet on these synthetic images and test on the real test set for both datasets.

\begin{table}
  \caption{The UViT performance.}
  \label{uvittraining}
  \centering
  \resizebox{0.8\columnwidth}{!}{%
  \begin{tabular}{lcccc}
    \toprule
     & FID & Sampler & Sampling Steps & Training Time \\
    \midrule
    CIFAR100 & 39.85 & DPM-Solver & 50 & \(\sim\) 27h \\
    TinyImageNet & 41.51 & DPM-Solver & 50 & \(\sim\) 54h \\
    \bottomrule
  \end{tabular}
  }
\end{table}

\subsection{Results}

\begin{figure}[tb]
   \centering
   \includegraphics[width=0.8\linewidth]{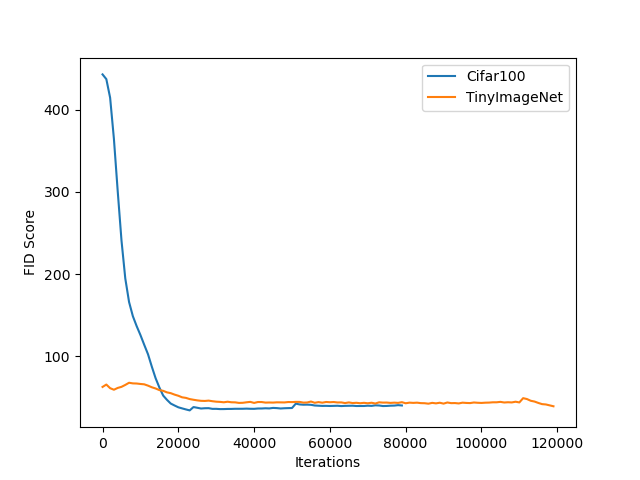}
   \caption{The FID scores in every \(1000\) iterations of training.}
   \label{fig:fidboth}
\end{figure}

We train the diffusion models UViT on CIFAR100 and TinyImageNet datasets respectively and the performances are shown in Table~\ref{uvittraining}. We use FID as the evaluation metric for synthetic image quality and present the FID scores at every \(1000\) training iterations in Fig~\ref{fig:fidboth}.

\begin{figure}[tb]
   \centering
   \includegraphics[width=\linewidth]{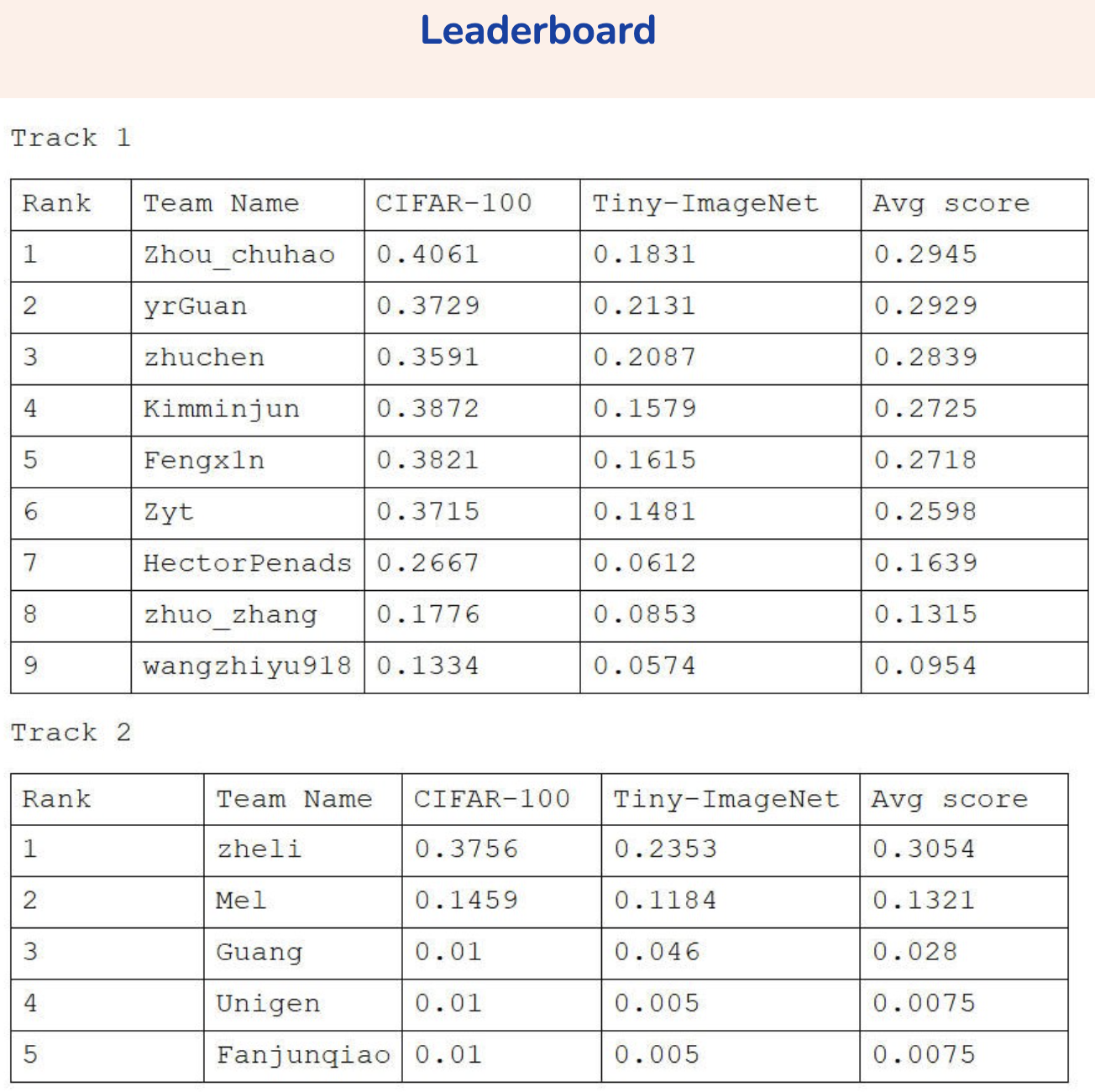}
   \caption{The Leaderboard of the Challenge. We are the rank \(1\) \textit{zheli} team in Track \(2\).}
   \label{fig:leaderboard}
   \vspace{-3mm}
\end{figure}

We compare our test accuracy in downstream classification task with other groups in the Track \(2\) of the Challenge as shown in Fig~\ref{fig:leaderboard}.
Our results are the \textit{Rank 1} \textit{zheli} team in the Track 2.

\begin{figure}[tb]
   \centering
   \includegraphics[width=\linewidth]{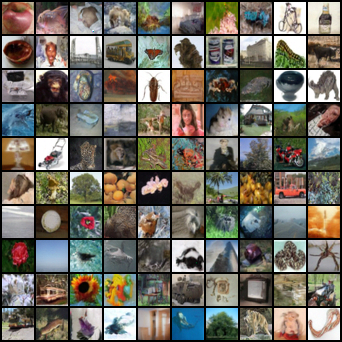}
   \caption{The generated images (\(32\times 32\)) by the diffusion model trained on CIFAR100 dataset. Each represents a class.}
   \label{fig:qualicifar}
   \vspace{-3mm}
\end{figure}

\begin{figure}[tb]
   \centering
   \includegraphics[width=\linewidth]{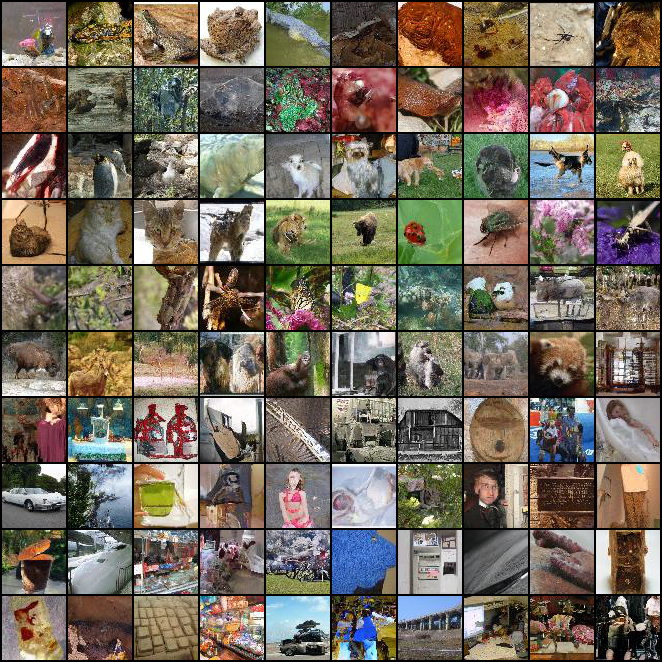}
   \caption{The generated images (\(64\times 64\)) from class 0 to 99 by the diffusion model trained on TinyImageNet dataset. Each represents a class.}
   \label{fig:qualitiny100}
   \vspace{-3mm}
\end{figure}

\begin{figure}[tb]
   \centering
   \includegraphics[width=\linewidth]{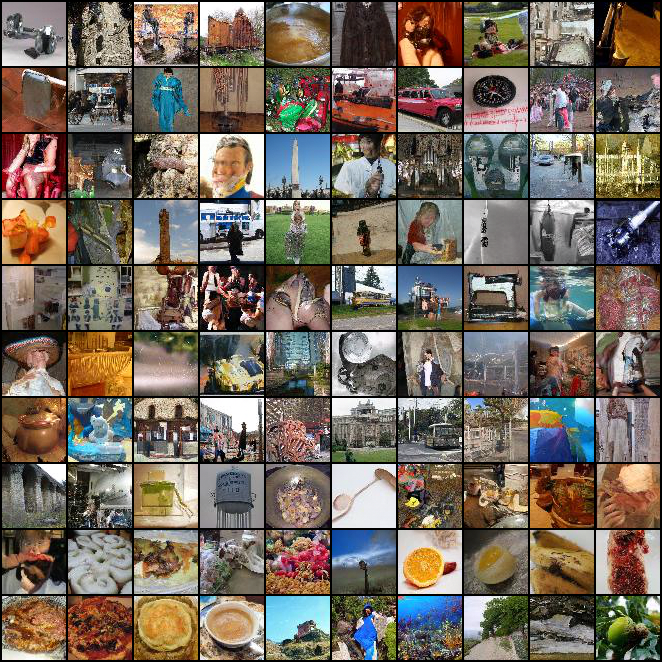}
   \caption{The generated images (\(64\times 64\))  from class 100 to 199 by the diffusion model trained on TinyImageNet dataset. Each represents a class.}
   \label{fig:qualitiny200}
   \vspace{-3mm}
\end{figure}

The generated images from CIFAR100 dataset are shown in Fig~\ref{fig:qualicifar}. Each image represents one class. Fig~\ref{fig:qualitiny100} presents the synthetic images of the first \(100\) classes from TinyImageNet dataset, while Fig~\ref{fig:qualitiny200} shows the remaining \(100\) classes.

\section{Conclusion}
In this paper, we propose a novel approach that leverages a diffusion model to generate distilled synthetic images representative of a large-scale image dataset. This can save data storage and computation cost as well as protect the privacy information.
The use of diffusion models enable the generation of an unlimited number of human-readable images to comprehensively represent the underlying data distribution. After generating images, we train classifiers on this small synthetic dataset, and report the test accuracy. This project ranked \(1\) in the Track \(2\) of The First Dataset Distillation Challenge at ECCV 2024.

\vspace{5mm}

\noindent\textbf{Acknowledgements:} This work was supported by the State of Bavaria, the High-Tech Agenda (HTA) Bavaria and HPC resources provided by the Erlangen National High Performance Computing Center (NHR@FAU) of the Friedrich-Alexander-Universität Erlangen-Nürnberg (FAU) under the NHR project b180dc. NHR@FAU hardware is partially funded by the German Research Foundation (DFG) - 440719683. Support was also received from the ERC - project MIA-NORMAL 101083647 and DFG KA 5801/2-1, INST 90/1351-1. WZ is supported by the JADS programme at Imperial College London and WS and SC by the UK Research and Innovation -- UKRI Centre for Doctoral Training in AI for Healthcare EP/S023283/1.

\clearpage  

%
%
\bibliographystyle{splncs04}
\bibliography{Paper-W498}

\end{document}